\documentclass[nohyperref]{article}

\usepackage{microtype}
\usepackage{graphicx}
\usepackage{booktabs} % for professional tables

\usepackage{hyperref}
 \usepackage[accepted]{icml2022}

\usepackage{amsmath}
\usepackage{amssymb}
\usepackage{mathtools}
\usepackage{amsthm}

\usepackage[capitalize,noabbrev]{cleveref}

\usepackage{tikz}
\usetikzlibrary{fit}					% fitting shapes to coordinates
\usetikzlibrary{backgrounds}	% drawing the background after the foreground
\usetikzlibrary{shapes.geometric} % draw the shapes
\usetikzlibrary{positioning}    % Relative Position
\usepackage{graphicx}
\usepackage{subcaption}
\usepackage{wrapfig,lipsum,booktabs}
\usepackage[format=plain,
            labelfont=it]{caption}
\usepackage{enumitem}

\theoremstyle{plain}
\newtheorem{theorem}{Theorem}[section]

\theoremstyle{definition}
\newtheorem{definition}[theorem]{Definition}

\theoremstyle{remark}

\usepackage[textsize=tiny]{todonotes}

\icmltitlerunning{Variational Mixtures of ODEs for Inferring Cellular Gene Expression Dynamics}

\begin{document}

\twocolumn[
\icmltitle{Variational Mixtures of ODEs for Inferring Cellular Gene Expression Dynamics}



\icmlsetsymbol{equal}{*}

\begin{icmlauthorlist}
\icmlauthor{Yichen Gu}{umeecs}
\icmlauthor{David Blaauw}{equal,umeecs}
\icmlauthor{Joshua Welch}{equal,umeecs,umcmb}
\end{icmlauthorlist}

\icmlaffiliation{umeecs}{Department of Electrical Engineering and Computer Science, University of Michigan, Ann Arbor, United States}
\icmlaffiliation{umcmb}{Department of Computational Medicine and Bioinformatics, University of Michigan, Ann Arbor, United States}

\icmlcorrespondingauthor{David Blaauw}{blaauw@umich.edu}
\icmlcorrespondingauthor{Joshua Welch}{welchjd@umich.edu}

\icmlkeywords{Variational Autoencoder, RNA Velocity, Generative Model}

\vskip 0.3in
]



\printAffiliationsAndNotice{\icmlEqualContribution} % otherwise use the standard text.

\begin{abstract}
A key problem in computational biology is discovering the gene expression changes that regulate cell fate transitions, in which one cell type turns into another. However, each individual cell cannot be tracked longitudinally, and cells at the same point in real time may be at different stages of the transition process. This can be viewed as a problem of learning the behavior of a dynamical system from observations whose times are unknown. Additionally, a single progenitor cell type often bifurcates into multiple child cell types, further complicating the problem of modeling the dynamics. To address this problem, we developed an approach called variational mixtures of ordinary differential equations. By using a simple family of ODEs informed by the biochemistry of gene expression to constrain the likelihood of a deep generative model, we can simultaneously infer the latent time and latent state of each cell and predict its future gene expression state. The model can be interpreted as a mixture of ODEs whose parameters vary continuously across a latent space of cell states. Our approach dramatically improves data fit, latent time inference, and future cell state estimation of single-cell gene expression data compared to previous approaches.
\end{abstract}

\section{Introduction}
\label{intro}

The human body contains many cell types with distinct forms and functions, which arise from progenitor cells in a stepwise developmental process. A key question in molecular biology is what regulates this process of cellular development. In general, the diversity of cell types arises not from cell-to-cell differences in the DNA sequence itself, but in which portions of the DNA sequence (genes) are used (expressed) in each cell. The central dogma of molecular biology states that genes are first transcribed into messenger RNAs (mRNAs) and these mRNAs are then translated into proteins, which carry out biochemical functions. The expression level of a gene in a cell can thus be quantified by the number of mRNA molecules present in the cell. Therefore, understanding cellular development requires modeling how mRNA expression changes over time. Such models are crucial for numerous areas of biology and medicine, such as neuroscience, cancer research, and regenerative stem-cell therapies.

We are interested in the following problem that arises in the context of modeling cellular gene expression changes. Each sample (cell), indexed by $i$, is represented by a vector $X_i(t) \in \mathbb{R}^d$ parametrized by time $t$. The trajectory $X_i(t)$ is governed by some differential equation plus random noise. However, for each $i$, only the vector $x_i := X_i(t_i)$ is observed at some unknown time $t_i$. Our goal is two-fold: recover the latent time $t_i$ for each sample and predict future states, i.e., $X_i(t)$ for $t > t_i$.

This unusual observation model stems from the limitations of single-cell RNA sequencing (scRNA-seq), the predominant experimental technology for measuring gene expression. The scRNA-seq technology ~\cite{tang2009mrna} counts the number of mRNA sequences expressed within a set of individual cells, ultimately yielding a matrix of expression levels for $~20,000$ genes across $10^4-10^6$ cells. However, measurement destroys the cell, so scRNA-seq gives only one single static snapshot of each cell at some moment in time. Second, the process of cell development is asynchronous--each cell takes a different amount of time to develop, so at a given moment, cells in a population will be at different developmental stages. An additional challenge is that a single starting cell type often bifurcates into multiple distinct cell types, so that cell-type-specific dynamics emerge over time. 

Our key insight is that knowledge about the biochemical steps required for gene expression can serve as a regularization or constraint for this otherwise ill-posed problem. By partially specifying the form of a differential equation describing the data generation process, we can simultaneously recover the unknown times and predict future states of the system. Our model can be interpreted as a variational autoencoder that reconstructs the data with an ODE whose parameters vary continuously across a latent space of cell states. Thus, we refer to our approach as a variational mixture of ODEs.

To our knowledge, this problem of learning a dynamical system from observations with unknown times has not been well studied. Previous papers have used both deep generative models and dynamical systems. But no previous work has demonstrated that these two approaches can be combined to solve the two-fold problem described earlier. Thus, this is a great example of an interesting  problem arising from a computational biology application. Our work also adds another item to the growing list of neural network models that achieve a new state of the art on a problem of high scientific interest. 

%Inspired by the ordinary differential equation (ODE) model in previous works~\cite{la2018rna,bergen2020generalizing}, we designed a variational autoencoder (VAE) based on the ODEs to solve the latent time inference problem. Furthermore, we extended a fixed ODE to a variational mixture of ODEs to model a continuous differentiation process. VAEs are very suitable for solving latent variable inference problems especially when we have no prior knowledge of what analytical form to choose. With the reparameterization trick, training is computationally efficient via gradient descent. 
The novel aspects of this work include:
\begin{enumerate}[leftmargin=*,nolistsep]
    \item We simultaneously estimate the times and dynamics of observations with unknown time labels. 
    \item By incorporating mechanistic insights about the biochemical process of gene expression, our model learns latent variables with clear biological meanings.
    \item Our approach dramatically improves the accuracy of time estimation and future state prediction compared to state-of-the-art approaches used by computational biologists, and thus has significant implications for biomedical research.
\end{enumerate}

%%%%%%%%%%%%%%%%%%%%%%%%%%%%%%%%%%%%%%%%%%%%%%%%%%%%%%%%%%%%%%%%%%%%%%%%%%%%%%%%%%%%%%%%%%%%%%%%
\section{Related Work}\label{relatedwork}
\textbf{Pseudotime Inference Methods.} Various methods have been applied to scRNA-seq data to uncover cellular development paths. Pseudotime inference methods use distance from a manually-specified starting cell to rank cells according to degree of development. Diffusion pseudotime~\cite{haghverdi2016diffusion} models cell development as a Markov process with a transition matrix. Other works~\cite{qiu2017reversed,SCHIEBINGER2019928} directly aim at determining the trajectory, i.e. putting the cells on one or multiple developmental paths. %Monocle~\cite{qiu2017reversed} applies reversed graph embedding to learn a low-dimensional embedding lying on a graph with tree structure. The tree structure reveals the relation between a stem cell type and multiple descendants. \citet{SCHIEBINGER2019928} applied optimal transport to study cell type transition in large datasets with multiple capture time points.

\textbf{RNA Velocity.} La Manno et al. \yrcite{la2018rna} developed the concept of RNA velocity based on the observation that both unspliced and spliced mRNA molecules appear in sequencing outputs. The relative ratio of spliced and unspliced counts can indicate whether the gene was being turned on or turned off at the time the cell was sequenced. They introduced an ODE model to describe the gene expression process and used a steady state assumption to estimate parameters. Later work \cite{bergen2020generalizing} relaxed the steady-state assumption, allowing all cells to be used in parameter estimation. These RNA velocity methods have been widely used by biologists to help understand cellular development processes \cite{plass2018cell,wilk2020single,litvivnukova2020cells} and are currently the state of the art in this area.

\textbf{Deep Generative Models for scRNA-seq Data.}
Previous papers have applied deep generative models to study scRNA-seq data. Many works~\cite{WANG2018320,lopez2018deep,gronbech2020scvae} have shown that variational autoencoders can learn useful latent representations for identifying cell types. In addition, \citet{lotfollahi2019scgen} showed that arithmetic operations of latent representation learned from scRNA-seq data can generate meaningful data corresponding to gene perturbation. BasisDeVAE~\cite{danks21basisdevae} used a VAE to simultaneously infer similarity-based pseudotime and cluster genes by their pseudotime trends. VeloAE~\cite{Qiaoe2105859118} embedded RNA velocity estimates from the steady-state model and spliced gene expression in the same latent space.

\textbf{Learning a Dynamical System.}
The problem of learning dynamical systems from high-dimensional datasets has been studied in many science and engineering domains. Early works~\cite{NIPS2008_07563a3f,pmlr-v31-dondelinger13a} applied gradient matching to estimate differential equation parameters. These methods involve MCMC sampling during the inference. Later works~\cite{NIPS2017_e71e5cd1,pmlr-v130-ghosh21b} improved the scalability and computational cost using variational inference. Another type of method called Neural ODEs \cite{chen2018,yildiz2019ode2vae,huang2020strode} was proposed to model time series. It assumes a dynamical system described by an ODE in the latent space. %A following work \cite{yildiz2019ode2vae} adds a second order ODE to describe the system. A recent work~\cite{huang2020strode} extends the neural ODE framework by modeling time as a temporal point process.

\textbf{Key Limitations of Previous Work.}
Each of these four classes of approaches has key limitations. Cell trajectory inference is based purely on pairwise similarity and cannot infer the directions or rates of cell development. RNA velocity enables mechanistic modeling of cell development, allowing quantitative analysis of gene expression and cell fate prediction. However, current methods have many limiting assumptions and fail to yield accurate results in many cases, such as when transcription rates vary over time or multiple lineages arise from the same progenitor cell type~\cite{bergen2021rna}. Deep generative models for single-cell data can learn cellular representations, but they have not incorporated the mechanistic insights from the RNA velocity approaches. General methods for learning dynamical systems require time information, so they are not directly applicable to datasets without time labels. To address these limitations, we propose VeloVAE, a variational mixture of ODEs that jointly recovers cell times and gene expression dynamics.

%%%%%%%%%%%%%%%%%%%%%%%%%%%%%%%%%%%%%%%%%%%%%%%%%%%%%%%%%%%%%%%%%%%%%%%%%%%%%%%%%%%%%%%%%%%%%%%%
\section{Methods}
\label{sec:method}
Section \ref{setup} and \ref{background} introduce the problem statement and background information about previous computational methods. Next, we describe a basic model that assumes fixed cellular dynamics to infer latent time in \ref{vae}. Finally, we describe our proposed method, a variational mixture of ODEs.
\subsection{Problem Setup}\label{setup}
\begin{figure}[t]
    \centering
    \includegraphics[width=0.9\columnwidth]{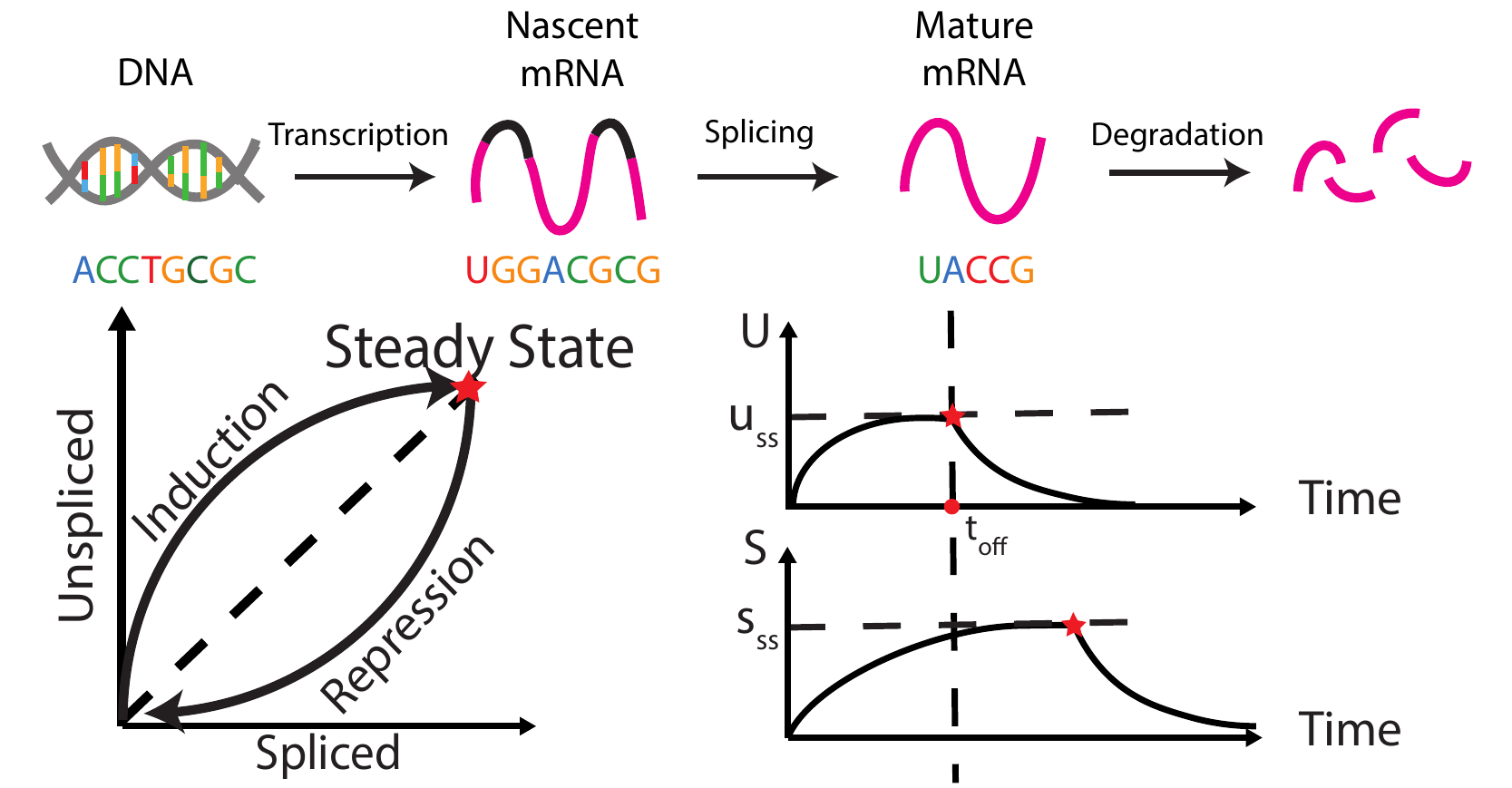}
    \caption{\textbf{Gene Expression Kinetics.} \emph{Top}: A gene is transcribed into nascent RNA before being spliced into mature RNA and subsequently degraded. \emph{Bottom}: temporal relationships between $u$ and $s$ implied by the model above. }
    \label{fig:rnavel}
\end{figure}

The key biochemical insight underlying our approach is that to express a gene, two types of RNA, nascent unspliced and mature spliced RNA, are produced sequentially. First, unspliced RNAs are directly transcribed from DNA sequences and contain non-protein-encoding sequences (introns). Next, the introns are removed so that nascent molecules are converted into mature ones (Fig. \ref{fig:rnavel}). To put it another way, increases in the unspliced count ($u$) for a gene must precede increases in the spliced count ($s$). This simple insight makes it possible to recover the ordering of cells lacking time labels. 

We assume that a dynamical system $F(t;\boldsymbol{\theta})$ generates scRNA count data. Here, $\boldsymbol{\theta}$ is a set of parameters describing the system, such as the transcription, splicing and degradation rates (introduced later; see Fig. \ref{fig:rnavel}). Our goal is to use observed scRNA data to simultaneously estimate the parameters $\boldsymbol{\theta}$ of $F$ and infer the unknown cell times $t$. 
\begin{definition}
Let $u_g$ and $s_g$ denote the unspliced and spliced mRNA count of the $g$-th gene. Let $\mathcal{G}=\{1,2,\ldots,G\}$ be a set of genes measured in an scRNA-seq experiment. The \textbf{feature vector} of a cell is defined as $\mathbf{x}=[u_1,u_2,\ldots,u_G,s_1,s_2\ldots,s_G]^T$.
\end{definition}
\begin{definition}
The \textbf{kinetic equation} of gene $g$ is defined as a system of ordinary differential equations relating changes in $u$ and $s$ over time. If there exists a solution $F(t;\boldsymbol{\theta})$ to the initial value problem with $u(0)=u_0, s(0)=s_0$, we call this solution the \textbf{kinetic function} for $g$.
\end{definition}
\begin{definition}
Given a kinetic function $u(t)$ and $s(t)$ of a gene, the \textbf{RNA velocity} of the gene is defined as $\frac{ds}{dt}$.
\end{definition}
\subsection{Modeling Gene Expression Kinetics}\label{background}
In previous work ~\cite{la2018rna}, the kinetic equation is modeled by a system of two linear ODEs:
\begin{align}
    \frac{du}{dt} = \alpha I_{\{t<t_{off}\}} - \beta u ,\;
    \frac{ds}{dt} = \beta u - \gamma s \label{eq:ode},
\end{align}
where $I_{\{\cdot\}}$ is an indicator function for the condition in brackets. The model parameters $\alpha,\beta$ and $\gamma$ correspond to the RNA transcription, splicing and degradation rates, respectively. The model assumes that two discrete phases can occur in the gene expression process: (1) induction, when new unspliced RNA molecules are being transcribed and (2) repression, when the transcription process stops and no new unspliced molecules are made. The induction phase is assumed to start at $t_{on}=0$ and the transition from induction to repression occurs at time $t_{off}$.

\textbf{Parameter Estimation by Steady-State Assumption.}
If the induction phase lasts for a long time, $u$ and $s$ will asymptotically converge to a stable value, called the steady state. We denote the steady-state values $u_{ss}$ and $s_{ss}$. The initial approach to estimating the parameters of the kinetic equation in the absence of cell times was to assume that the cells have reached steady state ~\cite{la2018rna}. A simple calculation shows that the steady-state condition of the kinetic equation \eqref{eq:ode} is $u_{ss}=\frac{\alpha}{\beta}$ and $s_{ss}=\frac{\alpha}{\gamma}$. Suppose we have a set of measurements of $u$ and $s$. We pick the top quantile, $u^*$ and $s^*$, as the approximate steady-state values. If we further assume that $\beta=1$,  then the estimated parameters are $\hat{\alpha}=u^*$, $\hat{\beta}=1$ (by assumption), and $\hat{\gamma} = \frac{u^*}{s^*}$.

\textbf{Dynamical Model and EM Algorithm.}
In practice, real-world datasets contain many cells that are not at the steady state; in fact, for some genes, only transient states are observed. The steady-state estimation method does not utilize these transient cells. Thus, \citet{bergen2020generalizing} developed a dynamical model called scVelo for estimating the parameters of the kinetic equation without the steady-state assumption. To deal with the absence of time, scVelo uses an expectation-maximization (EM) algorithm to jointly infer the latent times and model parameters.
In this approach, they first solve the kinetic equations \eqref{eq:ode} analytically to obtain the kinetic function:
\begin{align}
    u(t) &= u_0\exp(-\beta\tau) + \frac{\widetilde{\alpha}}{\beta}\left(1-\exp(-\beta\tau)\right) \label{eq:ut}\\
    s(t) &= s_0\exp(-\gamma\tau) + \frac{\widetilde{\alpha}}{\gamma}
    \left(1-\exp(-\gamma\tau)\right) \nonumber\\
    & +\frac{\widetilde{\alpha} - \beta u_0}{\gamma - \beta}\left(\exp(-\gamma\tau)-\exp(-\beta\tau)\right) \label{eq:st}\\
    \widetilde{\alpha} &:= \alpha I_{\{t<t_{off}\}},\;
    \tau := tI_{\{t<t_{off}\}} + (t-t_{off})I_{\{t\geq t_{off}\}}\nonumber
\end{align}
Note that the solution depends on the initial conditions $u(0)=u_0, s(0)=s_0$. ScVelo assumes that, given cell time $t$, $u$ and $s$ are Gaussian random variables whose means are given by the kinetic function \eqref{eq:ut},\eqref{eq:st}.  Because the dynamical model makes use of the full ODE solution, it does not require the steady-state assumption and produces better RNA velocity estimates.

\textbf{Limitations of scVelo.}
However, scVelo has several significant limitations. First, scVelo infers time separately for each gene, which neglects crucial information about the covariance of related genes and often leads to times that are inconsistent across genes. This gene-specific notion of time also makes it hard to compare the switch-off time (time when a cell stops producing new RNA) across genes. The lack of a common time scale, combined with the assumption that induction starts at $t=0$, also leads to frequent errors in estimating the overall direction of a gene (increasing or decreasing). Genes with a short or missing induction phase are particularly prone to being fit incorrectly by scVelo. Second, scVelo assumes a constant transcription rate $\alpha$ within the induction phase for each gene. In a recent review paper, the scVelo developers note that this assumption is often violated in real-world datasets, which leads to a variety of pathological behaviors ~\cite{bergen2021rna}. Finally, scVelo's model does not account for cell type bifurcations, which frequently occur in cellular development ~\cite{bergen2021rna}. 

\comment{\subsubsection{Variational Auto-Encoder (VAE)}
Variational auto-encoder~\cite{kingma2013auto} is a deep generative model used for learning low-dimensional representations. Given observation $\mathbf{x}\in\mathbb{R}^n$, the model assumes that some low-dimensional latent variable $\mathbf{z}\in\mathbb{R}^d$ that generates the data. Another perspective is that the high-dimensional data lie on a low-dimensional manifold. In many cases, the marginal likelihood $p(\mathbf{x})=\int p(\mathbf{x}|\mathbf{z})p(\mathbf{z})d\mathbf{z}$ is intractable so that the posterior $p(\mathbf{z}|\mathbf{x})=\frac{p(\mathbf{x}|\mathbf{z})p(\mathbf{z})}{p(\mathbf{x})}$ is also intractable. Thus, neither EM algorithms nor mean-field variational inference methods are applicable. To overcome this problem, \citet{kingma2013auto} developed the auto-encoding Variational Bayes algorithm. It picks a tractable approximate posterior $q_{\boldsymbol{\phi}}(\mathbf{z}|\mathbf{x})$ (encoder). The VAE model is obtained by using a neural network to parameterize $q_{\boldsymbol{\phi}}(\mathbf{z}|\mathbf{x})$ and the likelihood $p_{\boldsymbol{\theta}}(\mathbf{x}|\mathbf{z})$. Given a set of obervations $\mathbf{\mathcal{X}}=\{\mathbf{x_1},\mathbf{x_2},\ldots,\mathbf{x_n}\}$, the data likelihood $\log p(\mathbf{\mathcal{X}})$ is bounded by 
$$
    \mathcal{L}(\boldsymbol{\phi},\boldsymbol{\theta};\mathbf{\mathcal{X}}) = \sum_{i=1}^{n}\mathbb{E}_{q(\mathbf{z}|\mathbf{x_i})}\left[\log p(\mathbf{x_i}|\mathbf{z})\right] - KL\left(q(\mathbf{z}|\mathbf{x_i})\|p(\mathbf{z})\right)
$$
This is called the evidence lower bound (ELBO). By using the reparameterization trick, the gradient of ELBO with respect to the encoder parameters $\boldsymbol{\phi}$ can be computed using a batch mean, which allows fast and efficient computation.}

\subsection{VeloVAE: Basic Model (Fixed Transcription Rate)}\label{vae}
We first describe a deep generative model that recovers gene expression dynamics and cell time jointly assuming a single constant transcription rate for each gene. The model described in this section is thus a basic form of the variational mixture of ODE approach in section \ref{vaepp}. 

\textbf{Generative Process.} \label{gen_model}
We assume that cell time $t$ is first randomly sampled from a normal prior $\mathcal{N}(t_0, \sigma_0^2)$. If cell capture times are available (e.g., if cells were isolated separately on days 7 and 14), we can use them as an informative prior; otherwise, we can simply use a standard normal prior. We model the $u$ and $s$ counts for each gene using the kinetic function given by the analytical ODE solution, assuming that the genes are conditionally independent given cell times. Then, given $N$ i.i.d. time samples $t_1,t_2,\ldots,t_N$, gene expression data $\mathbf{x}_i$, $i=1,2,\ldots,N$, are generated by $\mathbf{x_i}=F(t_i;\boldsymbol{\theta})+\mathbf{r_i}, \mathbf{r_i}\sim \mathcal{N}(\mathbf{0},\boldsymbol{\Sigma_r})$. Here, $F(t;\boldsymbol{\theta})=[u_1(t),u_2(t),\ldots,u_G(t), s_1(t),s_2(t),\ldots,s_G(t)]^T$ is a vector containing all kinetic functions evaluated at $t$ and $\mathbf{r}$ is Gaussian random noise.  Equivalently, this means that the distribution of $\mathbf{x}$ conditioned on time is $p(\mathbf{x}|t,\boldsymbol{\theta})\sim \mathcal{N}(F(t;\boldsymbol{\theta}), \boldsymbol{\Sigma_r})$. We further assume that the noise variables $\mathbf{r}$ are mutually independent, i.e. $\mathbf{\Sigma_r}$ is diagonal with nonzero entries $\sigma_{u,1}^2,\sigma_{u,2}^2,\ldots,\sigma_{u,G}^2, \sigma_{s,1}^2,\sigma_{s,2}^2,\ldots,\sigma_{s,G}^2$. 

\textbf{ODE Formulation. }
We use a similar ODE model for $F(t;\boldsymbol{\theta})$ as in previous work~\cite{la2018rna,bergen2020generalizing}, with a slight modification. Instead of assuming all genes start generating mRNA at $t=0$, we allow asynchronous generation by adding a gene-specific parameter, $t_{on}$. This small change, in addition to inferring latent time jointly across genes, should alleviate scVelo's difficulty in fitting genes with an absent or short induction phase. The kinetic function thus has the same form as equations \eqref{eq:ut} and \eqref{eq:st}, except that the definition of $\tau$ changes:
$$\tau := (t-t_{on})I_{\{t_{on} \leq t  < t_{off}\}} + (t-t_{off})I_{\{t \geq t_{off}\}}$$

\begin{figure}[t]
    % Observation ([U,S])
\tikzstyle{observation}=[circle,
                        thick,
                        minimum size=0.7cm,
                        draw=black,
                        fill=gray!30]

% Latent Variable
\tikzstyle{latent}=[circle,
                    thick,
                    minimum size=0.7cm,
                    draw=black]

% A Group of Trainable Parameters
\tikzstyle{encdec}=[rectangle,
                    thick,
                    minimum size=0.7cm,
                    draw=black]

% Auxiliary Variable
\tikzstyle{aux}=[circle,
                    thick,
                    minimum size=0.7cm,
                    draw=black]

%Background
\tikzstyle{background}=[rectangle,
                        draw=black,
                        inner sep=0.15cm,
                        rounded corners=5mm]

    \begin{tikzpicture}
        
        % nodes of the inference model
         \node[observation] (u) {$\mathbf{u}$};
         \node[observation, right=of u, xshift=-0.25cm] (s) {$\mathbf{s}$};
         \node[encdec, below=of u, xshift=0.75cm, yshift=0.17cm] (enc) {$h(\mathbf{u},\mathbf{s};\boldsymbol{\phi})$};
         \node[latent,below=of enc,xshift=-1.5cm, yshift=0.19cm] (mut) {$\mu_t$}; %
         \node[latent,below=of enc,xshift=-0.5cm, yshift=0.19cm] (sigmat) {$\sigma_t$}; %
         \node[latent,below=of enc,xshift=0.5cm, yshift=0.19cm] (muc) {$\boldsymbol{\mu_c}$}; %
         \node[latent,below=of enc,xshift=1.5cm, yshift=0.19cm] (sigmac) {$\boldsymbol{\sigma_c}$}; %
         \node[latent,below=of sigmat,xshift=-0.4cm, yshift=0.19cm] (t) {$t$}; %
         \node[latent,below=of muc,xshift=0.5cm,yshift=0.27cm] (c) {$\mathbf{c}$}; %
         \node[aux,below=of sigmat,xshift=0.55cm, yshift=0.19cm] (eps) {$\boldsymbol{\epsilon}$}; %
        
        \path[->,thick]
       (u) edge (enc)
       (s) edge (enc)
       (enc) edge (mut)
       (enc) edge (sigmat)
       (enc) edge (muc)
       (enc) edge (sigmac)
       (mut) edge[dashed] (t)
       (sigmat) edge[dashed] (t)
       (muc) edge[dashed] (c)
       (sigmac) edge[dashed] (c)
       (eps) edge[dashed] (t)
       (eps) edge[dashed] (c)
       ;
       
        % nodes of the generative model
        \node[latent, label=right:{\scalebox{0.8}{$\sim \mathcal{N}(\mathbf{0}, \mathbf{I})$}}, right=of s, xshift=0.7cm] (cgen) {$\mathbf{c}$};
         \node[encdec, below=of cgen, yshift=0.8cm] (decrho) {$g(\mathbf{c};\boldsymbol{\theta_\rho})$};
         \node[latent, below=of decrho, yshift=0.8cm] (rho) {$\boldsymbol{\rho}$};
         
         \node[latent, label=right:{\scalebox{0.8}{$ \sim \mathcal{N}\left(t_0, \sigma_0^2\right)$}}, below=of decrho, xshift=1.0cm, yshift=0.8cm] (tgen) {$t$};
         \node[encdec, below=of rho, xshift=1cm, yshift=0.8cm] (dec) {ODE Solution: $F(t;\boldsymbol{\theta})$};
         
         \node[latent, below=of dec, xshift=-0.5cm, yshift=0.8cm] (uhat) {$\mathbf{\hat{u}}$};
         \node[latent, below=of dec, xshift=0.5cm, yshift=0.8cm] (shat) {$\mathbf{\hat{s}}$};
         \node[latent, left=of uhat, xshift=0.75cm] (sigmau) {$\boldsymbol{\sigma_u}$};
         \node[latent, right=of shat, xshift=-0.75cm] (sigmas) {$\boldsymbol{\sigma_s}$};
         \node[observation, below=of uhat, yshift=0.8cm] (ugen) {$\mathbf{u}$};
         \node[observation, below=of shat, yshift=0.8cm] (sgen) {$\mathbf{s}$};
         
        \path[->,thick]
        (tgen) edge (dec)
        (cgen) edge (decrho)
        (decrho) edge (rho)
        (rho) edge (dec)
        (dec) edge (uhat)
        (dec) edge (shat)
        (uhat) edge (ugen)
        (sigmau) edge (ugen)
        (shat) edge (sgen)
        (sigmas) edge (sgen)
        ;
        \begin{pgfonlayer}{background}
            \node [background,
                        fit=(u) (s) (mut) (sigmac) (t) (c),
                        label=Inference Model] {};
        \end{pgfonlayer}
        
        \begin{pgfonlayer}{background}
            \node [background,
                        fit=(tgen) (cgen) (sigmau) (sigmas) (ugen) (sgen),
                        label=Generative Model] {};
        \end{pgfonlayer}
   
    \end{tikzpicture}
    \caption{\textbf{Graphical Model.} Observed variables are colored gray. Dashed arrows indicate sampling.}
    \label{fig:graphmodel}
\end{figure}
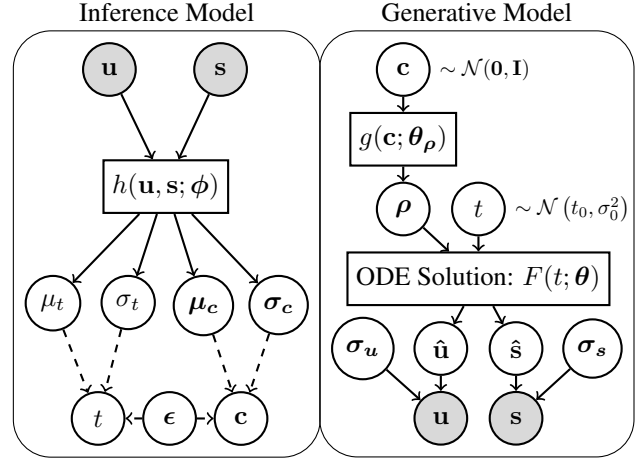

\textbf{Parameter Inference.}
Having formulated a generative model, our goal is to estimate both the ODE parameters $\boldsymbol{\theta}$ and the unknown cell times $t_i$. However, the posterior distribution of $t_i$ is intractable. Furthermore, unlike the scVelo model in which each gene has its own separate estimate of time, EM becomes much more difficult once cell time is shared across genes (see Appendix \ref{appdxA}). Instead, we use variational inference to find $t$ and $\boldsymbol{\theta}$. For our variational approximation, we use a Gaussian distribution whose parameters are output by a neural network. That is, $q(t|\mathbf{x})\sim\mathcal{N}(h(\mathbf{u,s};\boldsymbol{\phi}))$ where $h(\cdot)$ is a neural network that outputs mean and variance. Following the argument by~\citet{kingma2013auto}, we apply the reparameterization trick to approximate the evidence lower bound (ELBO) via sampling:
\begin{align}
    & ELBO = \sum_{i=1}^{N}\mathbb{E}_{q(t|\mathbf{x_i})}\left[\log p(\mathbf{x_i}|t)\right]-KL(q(t|\mathbf{x_i})||p(t))\nonumber\\
    \approx& \frac{1}{2}\sum_{i=1}^{N}\left[-2G\log(2\pi)-\log|\boldsymbol{\Sigma}_{\mathbf{r}}|-d(\mathbf{x_i},F(t_i;\boldsymbol{\theta});\boldsymbol{\Sigma}_\mathbf{r})^2\right]\nonumber\\
    & \quad +\frac{1}{2}\left[\log\frac{\sigma_p^2}{\sigma_q^2}+\frac{\sigma_q^2}{\sigma_p^2}+\frac{\left(\mu_p-\mu_q\right)^2}{\sigma_p^2}-1\right] \label{eq:elbo_velovae}
\end{align}
where $ F(t_i;\boldsymbol{\theta})$ is the kinetic function and $d(\cdot,\cdot;\boldsymbol{\Sigma})$ denotes the Mahalanobis distance with $\boldsymbol{\Sigma}$ as the covariance matrix. We can then jointly estimate the neural network weights $\boldsymbol{\phi}$, the ODE parameters $\boldsymbol{\theta}$, and the cell times $t_i$ by minimizing the negative ELBO using minibatch stochastic gradient descent. 

\textbf{Neural Network Architecture.}
The encoder is a multilayer perceptron (MLP) containing two hidden layers (500 and 250 neurons, respectively) with batch normalization~\cite{ioffe2015bn} and dropout~\cite{srivastava14dropout}. The bottleneck layer outputs the mean and standard deviation parameters of the variational distribution. 
%For the decoder, a common practice in VAE design is to use a neural network. Most previous VAE models on scRNA-seq data~\cite{lopez2018deep,WANG2018320,gronbech2020scvae,lotfollahi2019scgen,yu2021michigan} adopt this strategy to make use of the expressive power of neural networks. In contrast to these models, our VAE has a latent variable with very specific meaning: time. In doing so we resort to the domain knowledge and build the decoder directly from the ODE model (See section \ref{background}). Therefore, we make use of the kinetic function to transform the latent variable back to the feature vector, instead of a black-box neural network. We argue that this approach is useful for scientific discovery because both the latent variable and decoder parameters have physical meanings.

\begin{figure*}[h!]
    \centering
    \includegraphics[width=0.9\textwidth]{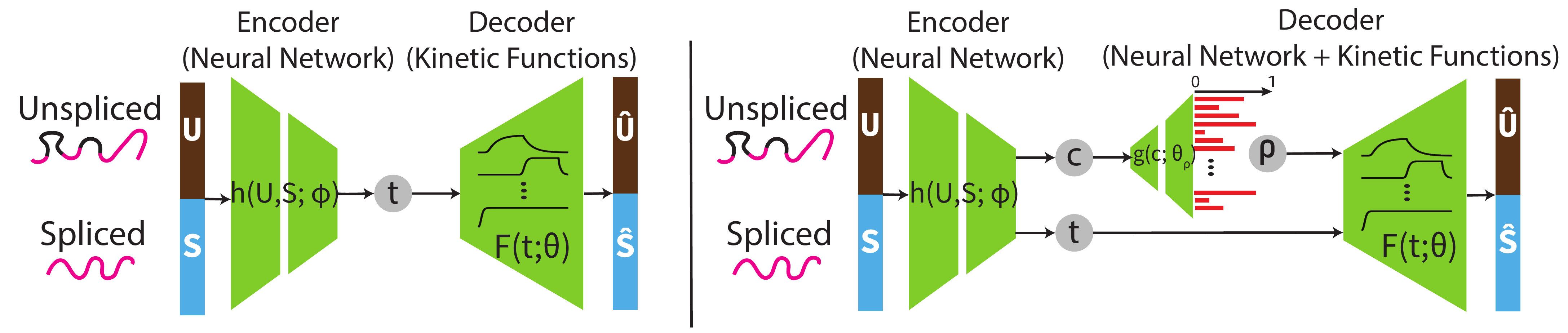}
    \label{fig:velovae}
    \caption{\textbf{VeloVAE Architecture.} \textbf{(a) Basic Model.} An encoder network infers a latent time from all $u$ and $s$ values for each cell. The data are reconstructed from inferred time using the kinetic function, whose analytical form is known. \textbf{(b) Full Model.} An encoder network infers both latent time and latent cell state $\mathbf{c}$. A decoder network generates the transcription rates $\boldsymbol{\rho}$, which are unique for each cell and each gene. The data are then reconstructed from $t$ and $\boldsymbol{\rho}$ using the kinetic function.}
\end{figure*}

\subsection{VeloVAE: Variational Mixture of ODE Model}\label{vaepp}
Although the model in the previous section can jointly infer cell times and ODE parameters, it still fails to capture important aspects of cellular development. In particular, constant transcription rates cannot account for bifurcations, which occur when a single type of stem cell develops into multiple descendant cell types. In fact, the possibility of bifurcations means that $u(t)$ and $s(t)$ may no longer be functions--multiple distinct cell states may be present at a given point in time. To capture these complex dynamics, we introduce a latent cell state variable $\mathbf{c}$ in addition to latent time and allow the transcription rate to vary smoothly over cell state space.

\textbf{ODE Formulation.} 
We adopt an ODE formulation similar to \eqref{eq:ode}, except that the transcription rate for each gene is not a single constant $\alpha$ anymore. Instead, we assume that the kinetic equation is a continuous mixture of ODEs with transcription rate parameters $\widetilde{\alpha} = \rho\alpha$. The relative transcription rate $\rho\in[0,1]$ is a function of latent cell state $\mathbf{c}$, and thus may be slightly different in each cell. The new kinetic equation is:
\abovedisplayskip=0pt
\belowdisplayskip=0pt
\begin{align}
    \frac{du}{dt} &= \rho\alpha - \beta u ,\quad
    \frac{ds}{dt} &= \beta u - \gamma s \label{eq:odes_proxy}
\end{align}

Note that there are no longer discrete induction and repression phases. This can be viewed as a generalization of \eqref{eq:ode}, since $\rho=1$ and $\rho=0$ correspond to the discrete induction and repression phases, respectively, used in the simpler formulation. Because $\rho$ is constant with respect to time, we can still solve the kinetic equation analytically to obtain a closed form for the kinetic function $F(t;\boldsymbol{\theta})$ in terms of $\alpha$, $\beta$, $\gamma$, and $\rho$. The solution is the same as \eqref{eq:ut} and \eqref{eq:st} except that $\widetilde{\alpha}=\rho\alpha$. Note also that for each gene, $\alpha$, $\beta$, and $\gamma$ are still shared across cells.
This model can now capture continuous transcription changes such as those in a bifurcating developmental process. 

\textbf{Generative Process.} \label{gen_vaepp}
%Inspired by Waddington's landscape~\cite{waddington2014strategy}, which likens cell development to rolling marbles on a landscape, we use a low-dimensional manifold to represent the landscape and assumes it contains the path of cell development. 
The generative process for the variational mixture of ODE model is as follows:
\begin{align*}
    t&\sim \mathcal{N}(t_0, \sigma_0^2),\; \mathbf{c} \sim \mathcal{N}(\mathbf{0,I})\\
    \boldsymbol{\tilde{\alpha}} &= \boldsymbol{\rho}\odot\boldsymbol{\alpha},\; \boldsymbol{\rho}=g(\mathbf{c};\boldsymbol{\theta_\rho})\\
    \mathbf{x}& \sim \mathcal{N}(F(t;\boldsymbol{\theta}),\boldsymbol{\Sigma}_{\mathbf{r}})
\end{align*}
Here, $g(\cdot)$ is a neural network with parameters $\boldsymbol{\theta_{\rho}}$, $\odot$ is the elementwise product, $F$ is the kinetic function of all genes, and $\boldsymbol{\Sigma}_{\mathbf{r}}$ is a diagonal covariance matrix. This generative process relies on a function $g$ mapping latent cell states $\mathbf{c}$ to relative transcription rates $\boldsymbol{\rho}$. Intuitively, the cell states can model continuous and bifurcating developmental paths, allowing the entire set of cells to be described as a family of ODEs whose parameters vary smoothly over the cell state manifold. We assume that $\boldsymbol{\rho}$ varies smoothly across the cell state space and that each point in cell state space maps to a unique $\boldsymbol{\rho}$. Although $g$ is deterministic for given $\mathbf{c}$, the inferred cell state for each cell $\mathbf{x}$ is probabilistic. Thus, the distribution of $\boldsymbol{\rho}$ can encompass multiple states near a bifurcation. Our generative model is summarized in figure \ref{fig:graphmodel}.

\textbf{Parameter Inference and Neural Network Architecture.}
The objective function is the ELBO shown in equation \eqref{eq:elbo_velovae}, with modified kinetic functions $F(t;\boldsymbol{\theta})$ and an updated KL divergence term incorporating the prior for $\mathbf{c}$. For $h$, we use the same MLP structure as the simple model with two additional outputs to produce the posterior mean and standard deviation of $\mathbf{c}$. We use an MLP that is the mirror image of $h$ (two layers with 250 and 500 neurons, respectively) to learn the mapping $g$ from $\mathbf{c}$ to $\boldsymbol{\rho}$. Source code is available online~\footnote{https://github.com/welch-lab/VeloVAE}.

\textbf{Initial Conditions.}
Because each cell now potentially has different ODE parameters, determining the initial conditions is more complex. Thus, instead of making the initial conditions trainable parameters, we simply train the model with $u_0=s_0=0$ in all of our experiments. This still yields excellent data reconstruction and latent time inference (Table \ref{tab:test_result}). However, the initial conditions are important for accurately predicting the future state of each cell. To improve the accuracy of future state prediction, we first train the VeloVAE to convergence using $u_0=s_0=0$ so that latent times and cell states are accurate, then determine the initial conditions for a cell at time $t$ by simply averaging the $(u,s)$ values observed in an immediately preceding time interval $[t-\delta_{1}, t-\delta_{2}]$. We then fine-tune the ODE parameters using these updated initial conditions, keeping latent time and cell state fixed.

\section{Experiments}\label{result}
\subsection{Datasets}
We evaluated our method on 6 different scRNA-seq datasets: pancreatic endocrinogenesis (PE)~\cite{bastidas2019pancreas}, dentate gyrus (DG1,DG2)~\cite{hochgerner2018dentategyrus,la2018rna}, embryonic E18 mouse brain cortex from 10X Genomics (MB1)\footnote{https://www.10xgenomics.com/resources/datasets/fresh-embryonic-e-18-mouse-brain-5-k-1-standard-1-0-0}, the erythroid lineage from mouse gastrulation (ET)~\cite{pijuan2019gastrulation}, and part of a whole mouse brain development dataset (MB2)~\cite{lamanno2021mousebrain}. See Appendix \ref{datasets} for details. Each dataset contains two cell-by-gene count matrices--one for unspliced counts and one for spliced counts. The matrices are preprocessed as described in the scVelo paper. %Briefly, genes with too few total mRNA counts or too little cell-to-cell variation are filtered out, generally leaving 1000-2000 genes. Next, a K-nearest neighbor graph is built based on principal components of the cells. Finally, unspliced and spliced counts are smoothed by averaging across the neighboring cells in the graph.
\subsection{Training}
For all experiments, we performed minibatch stochastic gradient descent using the ADAM optimizer with learning rate $2\times10^{-4}$ and batch size of 128. For each dataset, we trained on 70\% of the data until  the ELBO converged on the training set (number of epochs varied due to differences in dataset size), then evaluated the reconstruction error and likelihood on the held-out test set. We used 5 latent dimensions for cell state $\mathbf{c}$ in all experiments. For datasets with more than one capture time, we used the capture times to initialize the ODE parameters; otherwise, we used the steady-state approximation for initialization.
\subsection{Results}
We evaluated our method and compared it with scVelo, the state-of-the-art method for RNA velocity computation. To assess the importance of the mixture of ODEs, we also evaluated the basic model with fixed transcription rate.
We used several metrics to compare the performance of the methods. First, we assessed how well the models fit the observed data. The limitations of the single-cell data itself preclude ground truth for the cell times. However, the inferred times should at least be correlated with the cell capture times when available (usually on the order of days). We also evaluated the results qualitatively using biological knowledge of the overall properties of cellular development in the systems we studied. Our results show that VeloVAE fits the data and estimates cell times far more accurately than scVelo, while recovering qualitative properties of cellular development that scVelo cannot model. 

\begin{table}[h!]
\caption{Performance on scRNA-seq Datasets. We compare scVelo (SOTA), Basic Model (VAE with fixed rates), and VeloVAE (our proposed method). The metrics we use are (1) MSE = Mean Squared Error; (2) $k_t$ = Time correlation; and (3) $k_t$(Info.) = Time correlation under informative prior}
\label{tab:test_result}
%\vskip 0.15in
\begin{center}
\begin{small}
\begin{sc}
\begin{tabular}{lccccc}
\toprule
Dataset  & Method &  MSE &  $k_t$ & $k_t$(Info.)\\
\midrule
    & scVelo        & 2.107             & N/A   & N/A \\
PE  & Basic Model   & 6.815             & N/A   & N/A \\
    & VeloVAE       & \textbf{0.823}    & N/A   & N/A \\
\midrule
    & scVelo        & 0.670             & N/A   & N/A \\
DG1 & Basic Model   & 0.574             & N/A   & N/A \\
    & VeloVAE       & \textbf{0.243}    & N/A   & N/A \\
\midrule
    & scVelo        & 10.160            & N/A   & N/A\\
MB1 & Basic Model   & 10.431            & N/A   & N/A\\
    & VeloVAE       &\textbf{1.886}     & N/A   & N/A\\
\midrule
    & scVelo        & 0.873             & -0.707            & N/A \\
ET  & Basic Model   & 0.246             & \textbf{0.802}    & 0.802\\
    & VeloVAE       & \textbf{0.151}    & 0.622             & \textbf{0.855} \\
\midrule
    & scVelo        & 1.385             & -0.158            & N/A \\
DG2 & Basic Model   & 0.968             &  0.304            & 0.306\\
    & VeloVAE       & \textbf{0.159}    & \textbf{0.529}    &\textbf{0.707}\\
\midrule
    & scVelo        & 18.19             & -0.777            & N/A \\
MB2 & Basic Model   & 2.295             & 0.621             & 0.629 \\
    & VeloVAE       & \textbf{0.152}    & \textbf{0.870}    &\textbf{0.897} \\
\bottomrule
\end{tabular}
\end{sc}
\end{small}
\end{center}
\vskip -0.1in
\end{table}

\subsubsection{Data Reconstruction}
We used three metrics--mean squared error (MSE), mean absolute error (MAE), and log likelihood (LL)--to assess how well each method fits the data. For our two models, we calculated these metrics on both a training dataset (70\%) and held-out test dataset (30\%). Note that we are not able to calculate these metrics on a test set using scVelo, because it does not have a way to perform out-of-sample prediction. Training MSE results are shown in Table \ref{tab:test_result}; other metrics can be found in Appendix \ref{result_appdx}. The basic model generally achieves better MSE than scVelo, although the results are worse on the PE and MB1 datasets. This may be because scVelo fits each gene separately, estimating $N\times G$ latent time parameters (one for each cell and each gene) rather than $N$ latent time values estimated by the basic model. This allows scVelo to essentially overfit the data by separately adjusting the latent time values for each gene, but leads to severe inconsistency in cell time across genes and poor recovery of the overall cell times, as shown in Section \ref{results:time}. In contrast, the VeloVAE model consistently achieves the best reconstruction by a wide margin despite estimating only $N$ latent times. This suggests that the variational mixture of ODEs is crucial for accurately fitting the data. Furthermore, the test set is reconstructed nearly as accurately as the training set, indicating that the VeloVAE generalizes well and is not simply overfitting the training data.

\subsubsection{Time Inference}\label{results:time}
\begin{figure*}[h!]
    \centering
    \includegraphics[width=\textwidth]{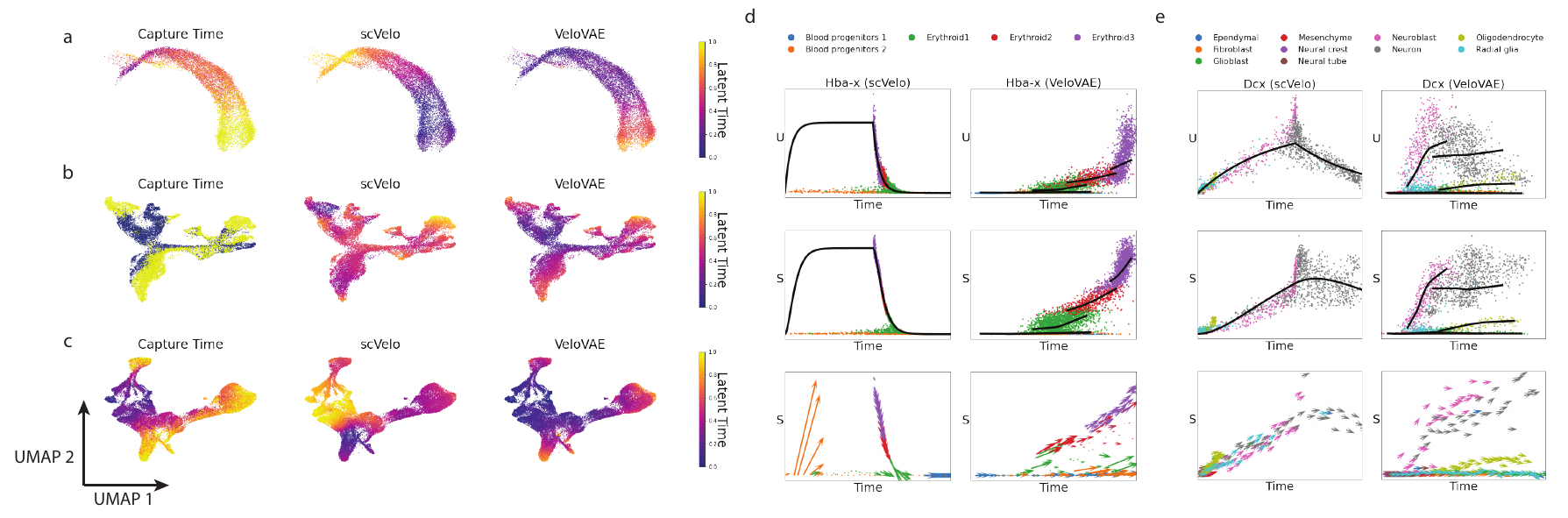}
    \caption{\textbf{Comparison of Inferred Time and Fit} \textbf{(a)-(c)} UMAP plots of scRNA data colored by capture time (left column), scVelo global time (middle), and VeloVAE time (right) for ET \textbf{(a)}, DG2 \textbf{(b)}, and MB2 \textbf{(c)} datasets. \textbf{(d)-(e)} Fitted (lines) and real (points) values from scVelo and VeloVAE for $\textit{Hba-x}$ gene in ET dataset and $\textit{Dcx}$ gene in MB2 dataset. Colors indicate cell types. Note that the VeloVAE fits are actually a point cloud, not a line (see inset plots in Fig. 5); the fit is so accurate that it would hide the real points, so we summarize the fit with a separate LOESS smooth per cell type to avoid overplotting. VeloVAE correctly models transcriptional boosts \textbf{(d)} and bifurcating gene expression trends \textbf{(e)}. Arrows in the bottom row of plots indicate predicted future cell states from RNA velocity estimates. Cells are randomly subsampled for clarity.}
    \label{fig:results}
\end{figure*}

Evaluating the latent time inference is challenging, because ground truth times are not available due to experimental limitations. However, three of the datasets (ET, DG2, MB2) contain data collected in multiple experiments across several days. The time stamps of these experiments (capture times) have very coarse granularity, and cells captured at the same time will span a wide range of developmental stages. Nevertheless, the inferred cell times should at least be correlated with the capture times. Thus, we computed the Spearman correlation between the cell times inferred by each method and the capture times.  %The global time estimation procedure involves multiple steps, including taking certain quantile of cell times across genes, using a heuristic to find starting cells, and smoothing times among cells with similar overall gene expression. 
\begin{wraptable}{r}{4.5cm}
\caption{Correlation between scVelo's gene-specific and global time, averaged across all genes}
\label{tab:scvcorr}
%\vskip 0.15in
\begin{center}
\begin{small}
\begin{sc}
\begin{tabular}{cc}
\toprule
Dataset & Correlation\\
\midrule
PE  & 0.262\\
DG1 & 0.097\\
MB1 & 0.226\\
ET  & 0.103\\
DG2 & -0.008\\
MB2 & -0.272\\

\bottomrule
\end{tabular}
\end{sc}
\end{small}
\end{center}
\vskip -0.1in
\end{wraptable}
Because VeloVAE can use the capture times as an informative prior for the cell times, we reported the correlation when using either a capture time prior or an uninformative prior in Table \ref{tab:test_result}. Although scVelo infers latent time separately for each gene, the tool provides a post-hoc procedure for estimating a single global time for each cell. Using this global time for comparison with our methods casts scVelo in the best possible light because the global time is more robust than the gene-specific latent times. Table \ref{tab:test_result} indicates that VeloVAE and the basic model both significantly outperform scVelo at inferring latent time. In fact, the scVelo global time is anticorrelated with capture time in all three datasets. In contrast, VeloVAE achieves the best performance, inferring latent times that are strongly correlated with capture time even with an uninformative time prior. The informative prior further increases the correlation. Figure \ref{fig:results} (a)-(c) visualize the true capture time and inferred cell time on the UMAP coordinates. 

The low time correlation from scVelo may be partly explained by inconsistency among the different notions of time fitted for each gene. To investigate this further, we computed the average time correlation between scVelo's gene-specific and global latent time. As Table \ref{tab:scvcorr} shows, the correlation is indeed quite low. Furthermore, it has been reported \cite{bergen2021rna} that genes whose kinetics violate some of the assumptions of scVelo's ODE model can lead to inferred time that proceeds in the wrong direction--consistent with what we observed here.

\subsubsection{Qualitative Advantages of VeloVAE}\label{result:kinetics}
\begin{figure}[h!]
    \centering
    \includegraphics[width=\columnwidth]{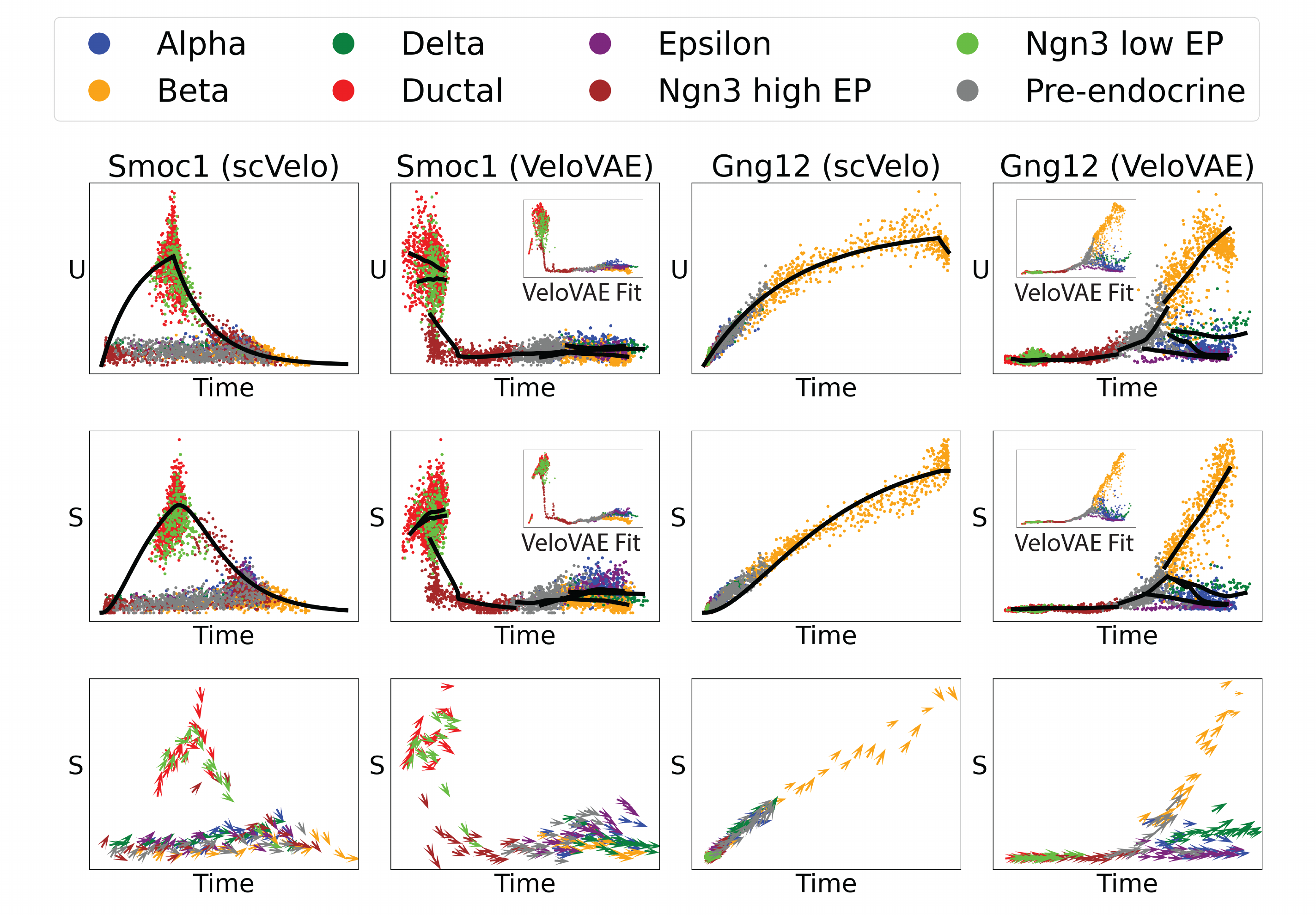}
    \caption{\textbf{VeloVAE Correctly Models Early Repression and Late Induction.} Fitted (lines) and real (points) values from scVelo and VeloVAE for the $\textit{Smoc1}$ (early repression) and $\textit{Gng12}$ (late induction, branching dynamics) genes in the PE dataset. Colors indicate cell types. Note that the VeloVAE fits are actually a point cloud, not a line; the fit is so accurate that it would hide the real points, so we summarize the fit with a separate LOESS smooth per cell type to avoid overplotting. The inset plots show the complete point clouds predicted by VeloVAE.}
    \label{fig:pancreas_sig}
\end{figure}

\textbf{VeloVAE Fits Early Repression and Late Induction Genes.} The restrictive assumptions of scVelo's ODE model, in concert with the separate inference of time for each gene, lead to very poor fits for many genes. In particular, scVelo suffers from systematic errors in genes that are turned off at the beginning of the process (early repression) or do not turn on until late in the process (late induction). For example, 
Fig. \ref{fig:pancreas_sig} shows the predicted values for \textit{Smoc1} (early repression gene) and \textit{Gng12} (late induction gene) in the PE dataset. In this dataset, the endocrine progenitor cells (Ngn3 low EP and Ductal) develop into four terminal cell types, alpha, beta, delta and epsilon. To fit \textit{Smoc1}, scVelo rearranges the cell times to force an induction phase, creating a biologically incorrect ordering where progenitor cells appear in the middle of time and incorrectly predicting an increase in gene expression at the beginning of time. Similarly, when fitting \textit{Gng12}, scVelo rearranges cell times to force all of the cells into the induction phase, leading to the incorrect prediction that \textit{Gng12} expression is constantly increasing. In contrast, VeloVAE fits the correct trends.

\textbf{VeloVAE Detects Transcriptional Boosts.} A recent review~\cite{bergen2021rna} showed that current RNA velocity approaches cannot account for ``transcriptional boosts". These occur when the transcription rate rapidly increases over time, making $u(t)$ and $s(t)$ concave upward. This confounds the assumptions of the simple ODE model, leading to a time estimate that is backward. However, as shown in Fig. \ref{fig:results}d, VeloVAE is able to accurately model such genes because the $\rho$ parameter varies by cell.

\begin{figure}[h!]
    \centering
    \includegraphics[width=\columnwidth]{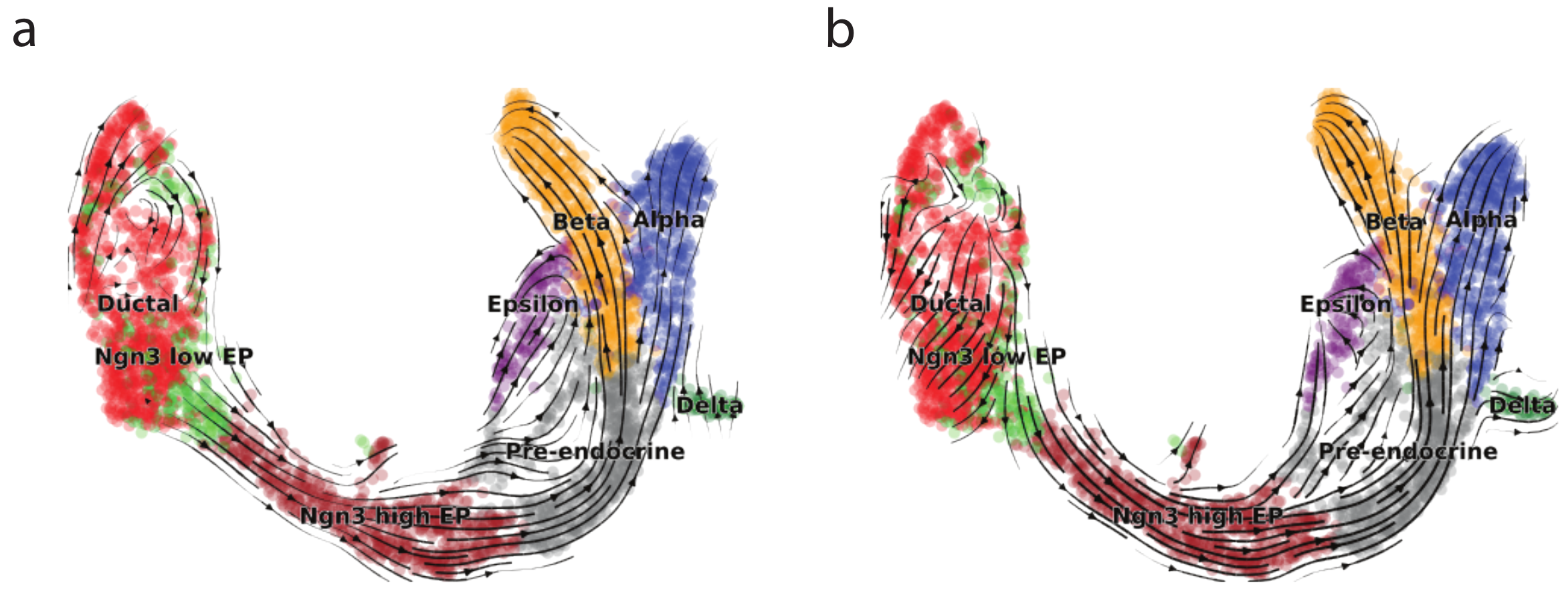}
    \caption{\textbf{Visualization of RNA Velocity from PE Dataset.} 2D projection of RNA velocity vectors predicted by scVelo \textbf{(a)} and VeloVAE \textbf{(b)}. VeloVAE more accurately predicts the branching dynamics to terminal cell types (alpha, beta, delta, and epsilon).}
    \label{fig:pancreas_vel}
\end{figure}

\textbf{VeloVAE Models Cell Type Bifurcations.} In most scRNA datasets (including 5 we analyzed here), a single progenitor type produces multiple cell types. A single ODE with a constant transcription rate cannot model time-varying kinetics, including bifurcation. Thus, neither scVelo nor our basic model can accurately model cell type bifurcations. However, VeloVAE flexibly models the emergence of cell-type-specific kinetics. For example, VeloVAE models the three-way branching expression pattern of \textit{Gng12}, which diverges as alpha, beta, and delta cells are formed (Fig. \ref{fig:pancreas_sig}). Similarly, VeloVAE models branching \textit{Dcx} expression in neurons and oligodendrocytes (Fig. \ref{fig:results}e). In contrast, scVelo rearranges gene-specific latent time to force the cells onto a single trajectory, erasing the cell-type-specific kinetics. A 2D visualization of RNA velocity for all genes in the PE dataset (Fig. \ref{fig:pancreas_vel}) confirms that VeloVAE better predicts branches leading to alpha, beta, delta, and epsilon cells.

\begin{figure}[h!]
    \centering
    \includegraphics[width=\columnwidth]{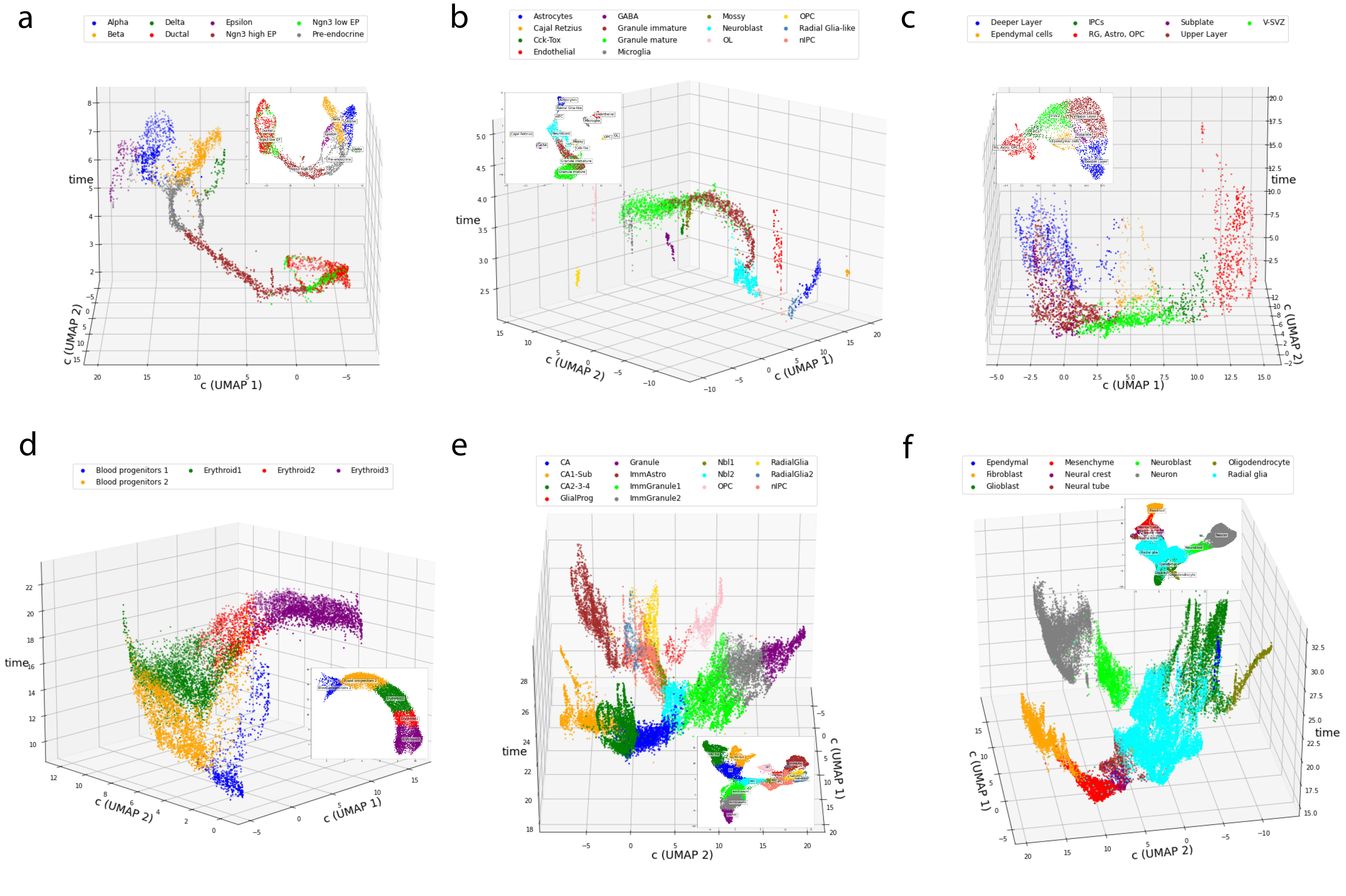}
    \caption{\textbf{Cell State Evolution over Time.} The vertical axis is the latent time and the horizontal plane contains 2D UMAP coordinates of $\mathbf{c}$. \textbf{(a)} Pancreas \textbf{(b)} Dentate Gyrus \textbf{(c)} 10x Mouse Brain \textbf{(d)} Erythroid \textbf{(e)} Dentate Gyrus 2 \textbf{(f)} Whole Mouse Brain. Insets show UMAPs from original expression data.}
    \label{fig:pancreas_cellstate}
\end{figure}

\textbf{Cell states are meaningful representations of cell differentiation.} In section \ref{sec:method}, we described the cell state as a continuous representation of cell types. We validate this claim by showing a 3D scatter plot of cell state versus time (Fig. \ref{fig:pancreas_cellstate}). For all six datasets, the cell state changes continuously over time and extends to multiple branches at cell type bifurcation points. 

\begin{figure}[h!]
    \centering
    \includegraphics[width=\columnwidth]{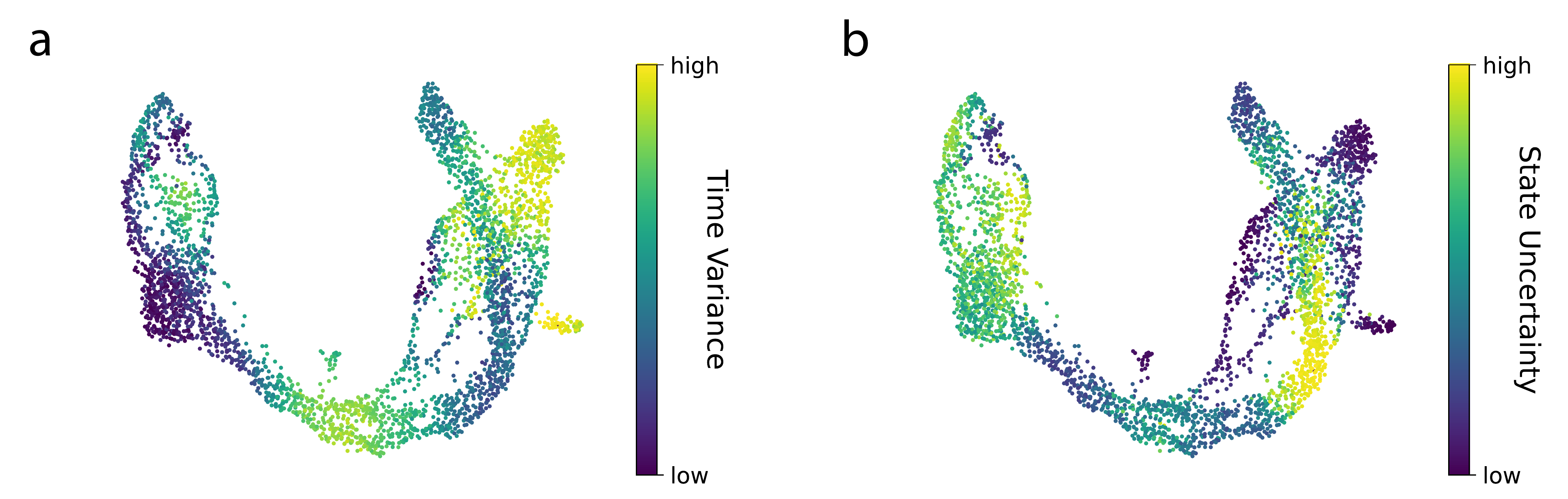}
    \caption{\textbf{Cell Time and State Uncertainty of the Pancreas Dataset.} We used CV as a measure of cell time and state uncertainty. The values are log-transformed for better visualization.}
    \label{fig:pancreas_variance}
\end{figure}

In addition, we note that the cell state models multiple states at bifurcation. We do not infer a single decision point in which a discrete cell fate decision occurs. Instead, we model the emergence of cell types as a smooth transition in which the cell state assignment has low uncertainty in undifferentiated progenitors, high uncertainty when cell fate decision is occurring, and low uncertainty again after the fate decision. To measure the uncertainty, we picked uni- and multi-variate coeffcient of variation (CV)~\cite{VANVALEN1974235} as our metric for uncertainty. For example, the CV of $\mathbf{c}$ is the highest for ductal cells deciding between cell cycle progression and exit to the Ngn3 progenitor state, as well as for Ngn3 progenitors deciding among the alpha, beta, delta, and epsilon fates (Figure \ref{fig:pancreas_variance}). 

\begin{figure}[h!]
    \centering
    \includegraphics[width=\columnwidth]{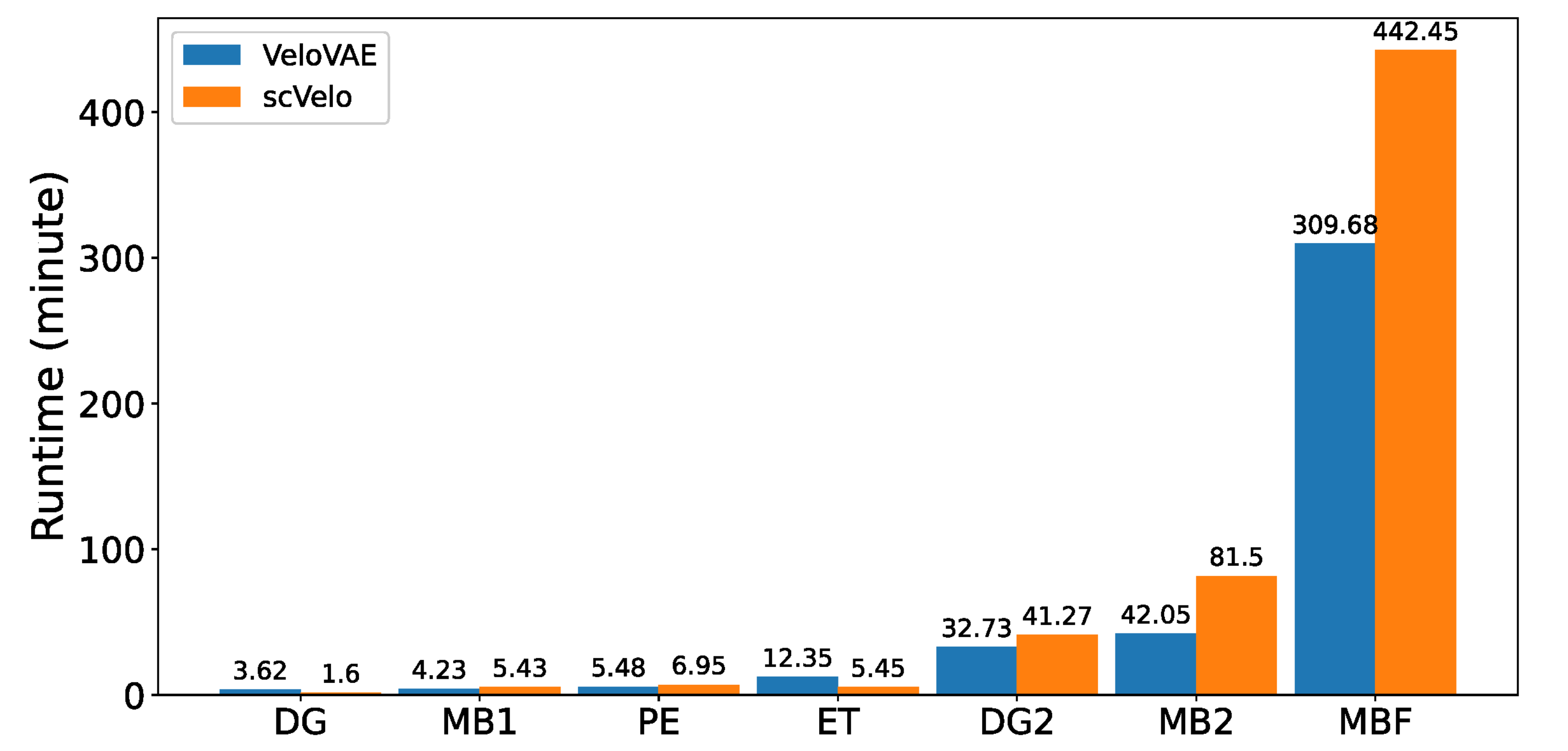}
    \caption{\textbf{Run-Time Comparison.} }
    \label{fig:runtime}
\end{figure}

\textbf{Scalability.} 
We think scalability is a key benefit of our approach. Minibatch optimization enables memory usage independent of cell number, whereas scVelo needs the entire dataset in memory. Number of iterations required by VeloVAE should also increase sublinearly with number of cells. As a rough benchmark, we trained our model for 600 epochs with an NVIDIA Tesla V100 GPU and ran scVelo on a single core of a 2.4 GHz Intel Xeon Gold 6148 CPU. We have not yet optimized our implementation for runtime or memory efficiency, and 600 epochs is likely overkill for large datasets. But we are already at least as fast as scVelo and 600 epochs on the whole mouse brain dataset took about 5 hours (Figure \ref{fig:runtime}).

\section{Discussion}
In this work, we developed VeloVAE, a deep generative model for inferring cellular gene expression dynamics. We demonstrated that VeloVAE can infer meaningful cell times while also fitting a model of gene expression dynamics. VeloVAE achieved much better performance than the state-of-the-art method, scVelo, on multiple scRNA-seq datasets.

Our principled probabilistic framework provides a strong foundation for future extensions. The current model assumes that genes are conditionally independent given both time and cell state. Relaxing this conditional independence assumption to infer groups of co-regulated genes is an exciting future direction. Another possible direction is modeling $u$ and $s$ as integer counts rather than normalized continuous variables.  

Our approach can be interpreted in several intuitive ways. From one perspective, we constrain the joint distribution of $u(t)$ and $s(t)$ to reflect our prior knowledge of the data generating process. From another point of view, our approach is a variational autoencoder modified so that the latent variables learned by the encoder have clear biological meanings (cell time and cell state) by construction. Another interpretation is that knowing any two of three quantities--time, observations, and underlying dynamics--enables inference of the third. Many previous papers have shown how to infer dynamics when time and observations are known; we show that having observations and general knowledge about how they are generated allows recovery of unknown times.

\nocite{langley00}

\section{Acknowledgements}
This work was funded by NIH grant R01HG010883 to JDW. We would like to thank Reetuparna Das for help with GPU resources, and Chen Li for helpful discussions.

\newpage
\bibliography{example_paper}
\bibliographystyle{icml2022}

%%%%%%%%%%%%%%%%%%%%%%%%%%%%%%%%%%%%%%%%%%%%%%%%%%%%%%%%%%%%%%%%%%%%%%%%%%%%%%%
%%%%%%%%%%%%%%%%%%%%%%%%%%%%%%%%%%%%%%%%%%%%%%%%%%%%%%%%%%%%%%%%%%%%%%%%%%%%%%%
% APPENDIX
%%%%%%%%%%%%%%%%%%%%%%%%%%%%%%%%%%%%%%%%%%%%%%%%%%%%%%%%%%%%%%%%%%%%%%%%%%%%%%%
%%%%%%%%%%%%%%%%%%%%%%%%%%%%%%%%%%%%%%%%%%%%%%%%%%%%%%%%%%%%%%%%%%%%%%%%%%%%%%%
\newpage
\appendix
\onecolumn
\section{Expectation-Maximization is Intractable When Cell Time is Shared Across Genes}\label{appdxA}
%\textbf{Dynamical Model of scVelo} The dynamical model~\cite{bergen2020generalizing}. assumes that, for each gene, there's a set of linear ordinary differential equations describing the splicing kinetics
%\begin{align*}
%    \frac{du}{dt} &= \alphaI_{\{t<t_{off}\}} - \beta u \\
%    \frac{ds}{dt} &= \beta u - \gamma s 
%\end{align*}
%Here, the parameters $\alpha,\beta$ and $\gamma$ are interpreted as the generation rate, splicing rate and degradation rate. They correspond to the biological process of transcribing new immature mRNA, converting immature RNA to mature RNA via splicing, and destruction of used RNA. $t_{off}$ is the switching point between induction and repression. The analytical solution to the ODE is
%\begin{align*}
%    u(t) &= u_0\exp(-\beta\tau) + \frac{\widetilde{\alpha}}{\beta}\left(1-\exp(-\beta\tau)\right)\\
%    s(t) &= s_0\exp(-\gamma\tau) + \frac{\widetilde{\alpha}}{\gamma}
%    \left(1-\exp(-\gamma\tau)\right)+\frac{\widetilde{\alpha} - \beta u_0}{\gamma - \beta}\left(\exp(-\gamma\tau)-\exp(-\beta\tau)\right) \\
%    \widetilde{\alpha} &:= \alpha I_{\{t<t_{off}\}} \\
%    \tau &:= tI_{\{t<t_{off}\}} + (t-t_{off})I_{\{t\geq t_{off}\}}
%\end{align*}
%Here, $\tau$ is the time interval length from the start of the phase $k$ to the cell time $t$. Both $k$ and $t$ are considered as latent variables. 

\textbf{Gene-Shared Latent Times.} It is shown in scVelo~\cite{bergen2020generalizing} that given the $u$ and $s$ values of a single gene, cell time can be inferred using an EM algorithm. However, their approach results in different cell times for different genes. Instead, we would like to develop an EM algorithm to infer the unique cell time, which is shared across all genes. We denote time as $\mathbf{t}$ and the ODE parameters as $\boldsymbol{\theta}$. Following the standard EM algorithm, we obtain the E-step at the $(j+1)$-th iteration:
\begin{align*}
    \mathcal{L}(\boldsymbol{\theta};\boldsymbol{\theta^{(j)}}) &= \mathbb{E}_{p(\mathbf{t}|\mathbf{X};\boldsymbol{\theta^{(j)}})}\left[\ln p(\mathbf{X}|\mathbf{t};\boldsymbol{\theta})\right] \\
    &= \sum_{i=1}^{N}\mathbb{E}_{p\left(t^{(i)}|\mathbf{x}^{(i)};\boldsymbol{\theta^{(j)}}\right)}\left[\ln p(\mathbf{x}^{(i)}|t^{(i)};\boldsymbol{\theta})\right]
\end{align*}
Here, we make the assumption that $t_i$ and $\mathbf{x_i}$ ($i=1,2,\ldots,N$) are mutually independent.
First, without computing the exact form, we can show that the posterior is intractable. 
\begin{align*}
    p(t|\mathbf{x};\boldsymbol{\theta}) &= \frac{p(\mathbf{x}|t;\boldsymbol{\theta})p(t)}{\int_{t'=-\infty}^{+\infty}p(\mathbf{x}|t';\boldsymbol{\theta})p(t')dt'}
\end{align*}
It's natural to assume that the time prior $p(t)$ is uniform in $[0,T]$. However, VeloVAE assumes a Gaussian prior $\mathcal{N}(\mu_0,\sigma_0^2)$ because the support of a uniform distribution is not $\mathbb{R}$ and the KL divergence might be undefined in some cases. For the purpose of analysis, we choose the uniform prior here. Later in the case of unshared latent time, we will see that with certain approximations, using a uniform prior results in the same algorithm as scVelo.

In addition, we assume the covariance matrix of $u$ and $s$ of all genes is diagonal, i.e. $\boldsymbol{\Sigma}=\mbox{diag}(\sigma_{u,1},\ldots,\sigma_{u,G},\sigma_{s,1},\ldots,\sigma_{s,G})$. Since $p(\mathbf{x}|t';\boldsymbol{\theta})$ is Gaussian, we have
\begin{align}
    p(t|\mathbf{x};\boldsymbol{\theta}) &= \frac{\frac{1}{(2\pi)^G|\boldsymbol{\Sigma}|^{\frac{1}{2}}}e^{-d(t)^2}\cdot \frac{1}{T}I_{\{t\in[0,T]\}}}{\int_{t'=-\infty}^{+\infty}\frac{1}{(2\pi)^G|\boldsymbol{\Sigma}|^{\frac{1}{2}}}e^{-d(t')^2}\cdot\frac{1}{T}I_{\{t'\in[0,T]\}}dt' } = \frac{e^{-d(t)^2}I_{\{t\in[0,T]\}}}{\int_{t'=0}^{T}e^{-d(t')^2}dt' }\\
    \mbox{where } d(t)^2 &:= \sum_{g=1}^{G}\frac{1}{2\sigma_{u,g}^2}(u_g-\hat{u}_g(t))^2+\frac{1}{2\sigma_{s,g}^2}(s_g-\hat{s}_g(t))^2 \label{eq:totalll}
\end{align}
Now consider the integral in the denominator. We know that both $\hat{u}(t)$ and $\hat{s}(t)$ have at least one exponential term involving $t$. Without exact calculation, we know that the denominator will involve the integral of $\exp(\exp(-ct))$ with some constant $c$ and this cannot be expressed by any elementary function. Let $C(\boldsymbol{\theta})$ be the constant equal to the integral in the denominator. Therefore, the total likelihood function is
\begin{align*}
    \mathcal{L}(\boldsymbol{\theta};\boldsymbol{\theta^{(j)}})
    = \sum_{i=1}^{N}\int_t\frac{e^{-d\left(t^{(i)};\boldsymbol{\theta^{(j)}}\right)^2}}{C^{(i)}(\boldsymbol{\theta^{(j)}})}\left[-G\ln(2\pi)-\frac{1}{2}\ln(|\boldsymbol{\Sigma}|)-d(t^{(i)};\boldsymbol{\theta})^2 \right]
\end{align*}
Because $C^{(i)}(\boldsymbol{\theta^{(j)}})$ is intractable, it's hard to directly optimize the total likelihood function. Hence, the EM algorithm cannot be easily applied. 

In addition, if we pick a Gaussian prior, we would end up with the same result with slight difference in the form of $d(t)$:
\begin{align*}
    d(t)^2 := \frac{(t-t_0)^2}{2\sigma_0^2}+\sum_{g=1}^{G}\frac{1}{2\sigma_{u,g}^2}(u_g-\hat{u}_g(t))^2+\frac{1}{2\sigma_{s,g}^2}(s_g-\hat{s}_g(t))^2
\end{align*}

\textbf{Unshared Latent Times.}
Now let's consider the special case of $G=1$, which is just the local gene fitting in scVelo. The M-step in scVelo~\cite{bergen2020generalizing} is simply to minimize the sample mean square error: 
\begin{align}
    \boldsymbol{\theta^{(j+1)}}&=\arg\min_{\boldsymbol{\theta}}\frac{1}{N}\sum_{i=1}^{N}\left[{\left(u^{(i)}-\hat{u}^{(i)}(t)\right)^2}+{\left(s^{(i)}-\hat{s}^{(i)}(t)\right)^2}\right] \label{eq:scv_em}
\end{align}
where $\hat{u},\hat{s}$ are predictions by the learned kinetic function of the gene. First, we need to assume $C^{(i)}(\boldsymbol{\theta^{(j)}})$ is a constant for all $i$. With this approximation, the M-step becomes
\begin{align}
    & \max_{\boldsymbol{\theta}}\mathcal{L}(\boldsymbol{\theta};\boldsymbol{\theta^{(j)}}) \nonumber\\
    = & \max_{\boldsymbol{\theta}}\sum_{i=1}^{N}\int_{t^{(i)}=0}^{T}\frac{e^{-d\left(t^{(i)};\boldsymbol{\theta^{(j)}}\right)^2}}{C^{(i)}(\boldsymbol{\theta^{(j)}})}\left[-c_\sigma-d(t^{(i)};\boldsymbol{\theta})^2 \right]dt^{(i)}\nonumber\\
    = & \min_{\boldsymbol{\theta}}\sum_{i=1}^{N}\int_{t^{(i)}=0}^{T}\frac{e^{-d(t^{(i)};\boldsymbol{\theta^{(j)}})^2}}{C^{(i)}(\boldsymbol{\theta^{(j)}})}d(t^{(i)};\boldsymbol{\theta})^2 dt^{(i)}\\
    \approx & \min_{\boldsymbol{\theta}}\frac{1}{C(\boldsymbol{\theta^{(j)}})}\sum_{i=1}^{N}\int_{t^{(i)}=0}^{T} d(t^{(i)};\boldsymbol{\theta})^2 e^{-d\left(t^{(i)};\boldsymbol{\theta^{(j)}}\right)^2}dt^{(i)} \label{eq:em}
\end{align}
We further assume $\sigma_u=\sigma_s=\sigma$, so $d(t)^2=\frac{1}{2\sigma^2}\left[(u-\hat{u}(t))^2+(s-\hat{s}(t))^2\right]$. We assume that $\sigma\approx0$ and $d(t;\boldsymbol{\theta})$ has a global minimum $t_0\in[0,T]$. As $\sigma$ approaches $0$, $d(t;\boldsymbol{\theta}^{(j)})$ approaches infinity. Following the analysis by \citet{li2020mathematics}, we apply the Laplace's method to approximate the integral:
\begin{align}
    \int_{t^{(i)}=0}^{T} d(t^{(i)};\boldsymbol{\theta})^2 e^{-d\left(t^{(i)};\boldsymbol{\theta^{(j)}}\right)^2}dt^{(i)} &\approx \sqrt{\frac{2\pi}{2d''(t_0;\boldsymbol{\theta^{(j)}})d(t_0;\boldsymbol{\theta^{(j)}})+2d'(t_0;\boldsymbol{\theta^{(j)}})^2}}e^{-d\left(t_0;\boldsymbol{\theta^{(j)}}\right)^2}d(t;\boldsymbol{\theta})^2 \\
    &\displaystyle \propto d(t;\boldsymbol{\theta})^2 \label{eq:laplace}
\end{align}
Using \eqref{eq:em} and \eqref{eq:laplace}, we obtain the final result:
$$
\arg\max_{\boldsymbol{\theta}}\mathcal{L}(\boldsymbol{\theta};\boldsymbol{\theta^{(j)}}) = \arg\min_{\boldsymbol{\theta}}\frac{1}{N}\sum_{i=1}^{N}\left[{\left(u^{(i)}-\hat{u}^{(i)}(t)\right)^2}+{\left(s^{(i)}-\hat{s}^{(i)}(t)\right)^2}\right]
$$
Therefore, minimizing the mean square error is equivalent to maximizing the total likelihood function under all the assumptions and approximations we made above.

\newpage
\section{Test Datasets}\label{datasets}
\begin{table}[tbh]
\caption{Dataset Description}
%\vskip 0.15in
\begin{center}
\begin{small}
\begin{sc}
\begin{tabular}{lcccc}
\toprule
Dataset Name & Cells & Genes & Cell Types & No. Time Points\\
\midrule
Pancreatic Endocrinogenesis (PE)  & 3696    & 2000  &   8   &  1\\
Dentate Gyrus (DG1)               & 2930    & 800   &   14  &  1\\
10x Mouse Brain (MB1)             & 3365    & 1000  &   7   &  1\\
Erythroid (ET)                    & 9815    & 1000  &   5   &  7\\
Dentate Gyrus (DG2)               & 18213   & 2000  &   14  &  2\\
Mouse Brain Development (MB2)*     & 29994   & 1000  &   10  &  20\\
\bottomrule
\end{tabular}
\end{sc}
\end{small}
\end{center}
\vskip -0.1in
\end{table}
*Subsampled to 30,000 cells

\section{Other Test Results}\label{result_appdx}
\begin{table*}[h!]
\caption{Performance on scRNA-seq Datasets. The metrics we compared, from left to right, are (1) Training and Testing Mean Squared Error; (2) Training and Testing Mean Absolute Error; and (3) Log Likelihood. Test metrics are not reported for scVelo because the method does not allow out-of-sample prediction.}
\label{tab:full_result}
%\vskip 0.15in
\begin{center}
\begin{small}
\begin{sc}
\begin{tabular}{lccccc}
\toprule
Dataset  & Method &  MSE (Train, Test) & MAE (Train, Test) & LL (Train, Test)\\
\midrule
    & scVelo        & 2.107,  N/A   & 0.423,  N/A   & -1702,  N/A     \\
PE  & Basic Model   & 6.815, 5.163  & 0.356, 0.351  & 271.71, 274.68  \\
    & VeloVAE       & \textbf{0.823}, \textbf{0.616} & \textbf{0.191},\textbf{0.192} & \textbf{727.42}, \textbf{717.20}  \\
\midrule
    & scVelo        & 0.670, N/A    &  0.316, N/A   & -2287, N/A      \\
DG1 & Basic Model   & 0.574, 0.560  &  0.302, 0.304 & 41.43, 41.96    \\
    & VeloVAE       & \textbf{0.243}, \textbf{0.253} & \textbf{0.190}, \textbf{0.194} & \textbf{237.63}, \textbf{234.57} \\
\midrule
    & scVelo        & 10.160, N/A       & 0.947, N/A    & -1779, N/A   \\
MB1 & Basic Model   & 10.431, 10.254    & 0.916, 0.921  & -456.07, -498.74  \\
    & VeloVAE       &\textbf{1.886}, \textbf{1.942}     & \textbf{0.392}, \textbf{0.398} & \textbf{440.61}, \textbf{440.65}  \\
\midrule
    & scVelo        & 0.873,N/A     & 0.456, N/A    &  -809.3, N/A  \\
ET  & Basic Model   & 0.246,0.246   & 0.251, 0.151  & 42.83, 44.25  \\
    & VeloVAE       & \textbf{0.151},\textbf{0.161} & \textbf{0.194}, \textbf{0.196}   & \textbf{67.36}, \textbf{66.88} \\
\midrule
    & scVelo        & 1.385, N/A    & 0.366, N/A     & -3513, N/A      \\
DG2 & Basic Model   & 0.968, 0.970  & 0.294, 0.295   & 954.74, 950.64  \\
    & VeloVAE       & \textbf{0.159}, \textbf{0.163}   & \textbf{0.120}, \textbf{0.121}    & \textbf{1797.30}, \textbf{1791.32} \\
\midrule
    & scVelo        & 18.19, N/A    & 0.47, N/A     & -7258, N/A \\
MB2 & Basic Model   & 2.295, 2.440  & 0.359, 0.364  &  -328.16, -338.89  \\
    & VeloVAE       & \textbf{0.152}, \textbf{0.147}        & \textbf{0.089}, \textbf{0.091}        & \textbf{926.98}, \textbf{924.48}\\
\bottomrule
\end{tabular}
\end{sc}
\end{small}
\end{center}
\vskip -0.1in
\end{table*}

%%%%%%%%%%%%%%%%%%%%%%%%%%%%%%%%%%%%%%%%%%%%%%%%%%%%%%%%%%%%%%%%%%%%%%%%%%%%%%%
%%%%%%%%%%%%%%%%%%%%%%%%%%%%%%%%%%%%%%%%%%%%%%%%%%%%%%%%%%%%%%%%%%%%%%%%%%%%%%%

\end{document}